\documentclass[letterpaper, 10 pt, conference]{ieeeconf}  

\IEEEoverridecommandlockouts                              

\usepackage[T1]{fontenc}	
\usepackage[utf8]{inputenc}
\usepackage{graphics} 
\usepackage{epsfig} 

\let\proof\relax

\let\labelindent\relax

\usepackage{amsmath,mathtools,amsfonts,amsthm} 
\usepackage{amssymb}
\usepackage{accents}
\usepackage{verbatim}
\usepackage{enumitem,kantlipsum}
\usepackage{color}
\usepackage{cite}
\usepackage{xcolor}
\usepackage{float}
\usepackage{dblfloatfix}

\newtheorem{thm}{Theorem}

\newtheorem{rem}{Remark}
\newtheorem{ass}{Assumption}
\newtheorem{problem}{Problem}
\DeclareMathOperator{\sgn}{sgn}

\usepackage{comment}
\usepackage{accents}
\usepackage{url}
\newcommand\munderbar[1]{%
  \underaccent{\bar}{#1}}

\title{\LARGE \bf
Kinodynamic Motion Planning via Funnel Control for Underactuated Unmanned Surface Vehicles
}

\author{Dženan Lapandić$^{1}$, Christos K. Verginis$^{2}$, Dimos V. Dimarogonas$^{1}$ and Bo Wahlberg$^{1}$
\thanks{$^{1}$Dženan Lapandić, Dimos V. Dimarogonas and Bo Wahlberg are with  Division of Decision and Control Systems, KTH Royal Institute of Technology, Stockholm, Sweden.   {\tt\small lapandic,dimos,bo@kth.se}}%
\thanks{$^{2}$Christos K. Verginis is with Division of Signals and Systems, Department of Electrical Engineering, Uppsala University,
Uppsala, Sweden. {\tt\small christos.verginis@angstrom.uu.se
}
         }%
}

\begin{document}

\maketitle
\thispagestyle{empty}
\pagestyle{empty}

\begin{abstract}
We develop an algorithm to control an underactuated unmanned surface vehicle (USV) using kinodynamic motion planning with funnel control (KDF). 
KDF has two key components: motion planning used to generate trajectories with respect to kinodynamic constraints, and funnel control, also referred to as prescribed performance control, which enables trajectory tracking in the presence of uncertain dynamics and disturbances. 
We extend prescribed performance control to address the challenges posed by underactuation and control-input saturation present on the USV. 
The proposed scheme guarantees stability under user-defined prescribed performance functions where model parameters and exogenous disturbances are unknown. 
Furthermore, we present an optimization problem to obtain smooth, collision-free trajectories while respecting kinodynamic constraints.
We deploy the algorithm on a USV and verify its efficiency in real-world open-water experiments.
\end{abstract}

\section{Introduction}

Developing algorithms that increase the level of autonomy of an unmanned vehicle has been a challenging problem that has captivated the attention of a broad research community. 
From aerial to surface vehicles and autonomous cars, the adopted algorithms possess features that provide a limited degree of autonomy. Thus, intelligent decision-making and full autonomy are still open problems \cite{liu2016unmanned,campbell2012review}.
Unmanned surface vehicles (USVs) are meant to operate in open waters and thus are subject to uncertain weather conditions and wave and wind disturbances. 
Additionally, the model dynamics might not be fully known, and parameters might change during the operation. 
Moreover, USVs can have motion limitations due to the underactuation and placement of the installed control thrusters. 
All this makes the control of a USV challenging, especially in scenarios where USVs must satisfy performance and safety specifications.
We consider the problem of motion planning and control of a surface vehicle (a boat) by splitting it into two layers: 
\textit{trajectory generation}, which generates a trajectory based on the user input and the environment characteristics, such as obstacles, and \textit{trajectory tracking} that tracks the trajectory within prespecified performance and is robust to the model uncertainties and disturbances.  

Motion planning is one of the fundamental problems in robotics. In general, its task is to move an agent or an object from a start to a goal location without colliding with environment obstacles \cite{lavalle2006planning}. 
Furthermore, kinodynamic sampling-based motion planning is a class of motion planning algorithms that impose the kinodynamic constraints of the system on the generated trajectories \cite{donald1993kinodynamic,sucan2009kinodynamic,karaman2010optimal}. 
The advantage of kinodynamic motion planning is that it can produce trajectories suitable for trajectory tracking even when the dynamics are not fully known. 
Namely, the kinodynamic motion planning problem generates a trajectory using kinematic constraints, such as avoiding obstacles, and dynamic constraints, such as bounds on velocity, acceleration, and jerk, without the requirement of knowledge of complete agent's dynamical equations \cite{donald1993kinodynamic}.
To that respect, B-splines \cite{deboor2001splines} have recently regained attention for motion primitives generation in fast and iterative schemes for quadrotors and other unmanned agents \cite{zhou2019robust,tordesillas2021mader,van2017distributed}. 
Their polyhedral representation properties enable simple collision checking, while their derivative properties are suitable for enforcing velocity and acceleration constraints on trajectories \cite{tordesillas2021mader}.

\begin{figure}
\centering
    \includegraphics[width=0.95\columnwidth]{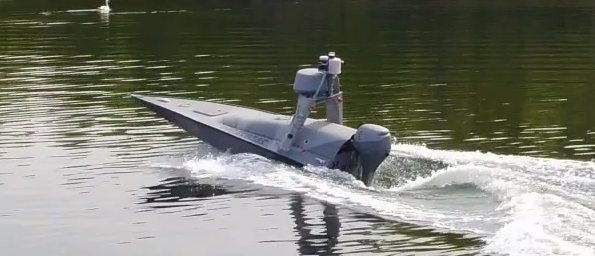} 
    \caption{Piraya autonomous unmanned surface vehicle. Courtesy of \protect\url{https://portal.waraps.org/} }
    
    \label{fig:piraya} \vspace{-0.5cm}
\end{figure}

The trajectory tracking problem for surface vehicles or vessels has been broadly studied in the last century \cite{fossen1994guidance}. 
For underactuated USVs (depicted in Fig.~\ref{fig:piraya}) characterized by \textit{three} degrees of freedom (position and orientation) and \textit{two} inputs, which we consider in this work, the classical approach is to steer the vessel along a path using the control of forward velocity and turning \cite{fossen2011handbook}. 
In \cite{aguiar2003position, aguiar2007trajectory}, the authors developed nonlinear Lyapunov-based controllers for path following and trajectory tracking for the considered underactuated vehicles with known or adaptively estimated model parameters. 
Another approach is introducing a frame change and designing a nonlinear controller in the new frame. The reference frame is usually chosen as the \textit{Serret-Frenet frame} and Lyapunov's direct method and backstepping ensure the path-following  \cite{skjetne2001nonlinear, do2004state}. 
However, this type of controller necessitates the assumption of a constant forward velocity. 
In more recent works \cite{pavlov2009mpc,alessandretti2013trajectory,abdelaal2018nonlinear}, the authors use MPC for trajectory tracking and autonomous docking of USVs \cite{martinsen2019autonomous}. However, most controllers in the literature assume accurate knowledge of the dynamical model and parameters. 

In this paper, we develop an algorithm for the kinodynamic motion planning problem of a USV. 
We extend the Kinodynamic motion planning via funnel control (KDF) framework developed in \cite{verginis2022kdf}.
First, we pose the motion planning problem as an optimization problem using B-splines to generate smooth kinodynamic trajectories satisfying both the spatial and dynamical constraints without using dynamical equations of motion. Second, we employ a parameter-ignorant control protocol able to bound the system behaviour to user-specified performance. To that end, we extend prescribed performance control (PPC) \cite{BECHLIOULIS20141217}  within the KDF framework to be applicable to the underactuated USV. 
PPC is a nonlinear control law that ensures all errors stay inside of user-defined prescribed performance functions or funnels \cite{bechlioulis2014low}. 
PPC, similar to funnel control \cite{ilchmann2002tracking,ilchmann1993non}, implicitly adapts its gains depending on how close the errors are to the funnels. 
However, it is important to highlight a specific limitation of the existing PPC methodology in the context of this work which is that it cannot be directly implemented to underactuated systems \cite{lapandic2022robust,verginis2022robust}.

Our main contributions are 1) an optimization-based algorithm that uses a sampling-based motion planner and generates kinodynamic collision-free trajectories, 2) appropriate modifications of the original PPC methodology to accommodate for the underactuated USV system, and 3) stability conditions for the closed-loop system under control input constraints.

Compared to \cite{verginis2022kdf}, first, we have extended the trajectory generation with an optimization problem which takes into account obstacles and constraints on velocity and acceleration for generating smooth B-splines. 
In \cite{verginis2022kdf}, a pure spline interpolation is done between the RRT obtained points that reside in the extended free space based on a fixed time interval constraint. 
This method could result in collisions with obstacles if the fixed time interval constraint between two points was not chosen adequately. 
Second, in this work, we consider a specific class of underactuated vehicles, necessitating a modification and extension of the KDF framework originally not applicable to underactuated nor non-holonomic robots. 
Third, we introduce input constraints due to the actuator saturation on the physical surface vehicle and provide conditions on the upper bounds of the dynamics and disturbances to ensure compliant behaviour.
Finally, we conducted the real-world, open-water experiments, significantly enhancing the realism and complexity compared to the previous paper with lab-based studies. This allows for a more robust assessment of the performance of our algorithm and system behaviour in unpredictable open-water environments.

It is worth noting that prescribed performance has been used for underwater underactuated vehicles in \cite{bechlioulis2016trajectory}. 
However, in our work, we develop a more general framework that includes motion planning, considers input constraints, and provides real-world experimental results. 

The paper is organized as follows. In Section II, we formulate the problem; in Section III, we provide preliminaries and the considered transformation of the dynamics. In Section IV, we propose a control design for the underactuated USV and present a stability result. The optimization-based kinodynamic motion planning is presented in Section V. In Section VI, we present real-world experimental results. Finally, Section VII concludes the paper.
\section{Problem Formulation}\label{sec:prob_statement}
We consider a USV with one rotating thruster at the rear. The model is the reduced 3-DoF (degrees of freedom) boat model, as described in \cite{fossen1994guidance}, with motion in surge, sway, and yaw: 
\begin{subequations}
\begin{align}
    \dot{\eta} &= R(\psi)\nu \\
    M\dot{\nu} + C(\nu)\nu + D(\nu)\nu &= \tau_{act} + \tau_d \label{eq:model_dyn}
\end{align}\label{eq:model}%
\end{subequations}
where $\eta=[p^T,\psi]$, $p\in \mathbb{R}^2$ is the position in the inertial frame, $\psi\in[0,2\pi)$ is the orientation of the boat in the inertial frame, and $\nu =[u,v,r]^T$ are forward velocity (surge), lateral velocity (sway) and angular velocity (yaw), respectively, expressed in the body frame. We further define $x=[\eta^T, \nu^T]^T \in \mathbb{X}$ as the state, where  $\mathbb{X}=\mathbb{R}^2\times[0,2\pi) \times \mathbb{R}^3$.  
Further, $R(\psi)\in SO(3)$ is the rotation matrix around the $z$-axis, $SO(3)$ is the special orthonormal group $SO(n) = \{ R \in \mathbb{R}^{n\times n} : RR^T = I_n, \det R = 1  \}$, $I_n \in \mathbb{R}^{n\times n}$ is the identity matrix, $M$ is the inertia matrix, $C(\nu)$ denotes the Coriolis and centripetal effects, and $D(\nu)$ is the drag matrix; 
$\tau_d(x,t)=[\tau_{d,x},\tau_{d,y},\tau_{d,\psi}]^T$ are unknown but bounded disturbances in $t$ and locally Lipschitz in $x$; $\tau_{act}=[X, Y, N]^T$ is the generalized control torque vector consisting of control forces $X$, $Y$ in forward and lateral directions, respectively, and the control torque $N$ for rotational motion, where the notation is adopted from the marine craft control theory. The dynamics in \eqref{eq:model_dyn} can be rewritten as

\begin{subequations}
\begin{align}
    m\dot{u}(t) &= f_u(x,t) + X \\
    m\dot{v}(t) &= -k_v v(t) + f_v(x,t) + Y \label{eq:2b}\\
    I_z\dot{r}(t) &= f_r(x,t) + N
\end{align}\label{eq:model_rewritten}%
\end{subequations}
where $m$ is the \textit{unknown} mass of the USV, $I_z$ is the \textit{unknown} moment of inertia around the $z$-axis and $f_u$, $f_v$, and $f_r$ are \textit{unknown} functions that model the drag, Coriolis and centripetal effects, as well as external disturbances. Note that, due to the conditions on the external disturbances $\tau_d$, these functions are locally Lipschitz in $x$ for each $t\in\mathbb{R}_{\geq0}$ as well as continuous and uniformly bounded in $t$ for each $x$. Also, in Eq.~\eqref{eq:2b}, we explicitly write the drag term for the unactuated part of the dynamics needed for the stability analysis in Sec.~\ref{ssec:stability}, where $k_v>0$ is a positive damping constant \cite{fossen2011handbook}.
The considered USV has one thruster at the rear, thus $\tau_{act}$ can further be defined as
\begin{equation}\label{eq:tau_act}
    \tau_{act} = \begin{bmatrix}
    X \\ Y \\ N
    \end{bmatrix} = \begin{bmatrix}
    F_T \cos(\alpha_r) \\
    F_T \sin(\alpha_r) \\
    \Delta_x F_T \sin(\alpha_r) 
    \end{bmatrix}
\end{equation}
where $F_T$ is the applied thrust and $\alpha_r$ is the applied rudder position representing the USV's control inputs, and $\Delta_x$ is the longitudinal displacement from the center of gravity. The underactuation stems from the fact that only two control inputs, $F_T$ and $\alpha_r$, are available for control of the 3-DoF system, and they all affect $X, Y$, and $N$. Moreover, two elements of $\tau_{act}$ are linearly dependent, i.e., $Y = c N$, $c\neq0$, thus further limiting the controlled behavior of the system.

Because of the engine's physical constraints, such as its limited thrust and inability to generate negative thrust, we assume control input constraints on the engine thrust $F_T \in [0,\bar{F}_T]$, for some positive $\bar{F}_T>0$. Moreover, the rudder position is physically constrained, and the angle is $\alpha_r \in [-\bar{\alpha}_r,\bar{\alpha}_r]$, where $\bar{\alpha}_r \in (0, \pi/6]$ represents the maximal rudder position.
Thus, the control inputs $F_T$ and $\alpha_r$ should satisfy the described control input constraints. 

Due to the aforementioned limitations in thrust, we impose restrictions on the attainable velocity and acceleration in the inertial frame. These constraints are articulated as $\| \dot{p}\|\leq \mathrm{v}_{\textup{max}}$, $\| \ddot{p}\|\leq \mathrm{a}_{\textup{max}}$, where $\mathrm{v}_{\textup{max}}$ and $\mathrm{a}_{\textup{max}}$ denote the maximum permissible velocity and acceleration respectively. These limitations substantially curtail the performance capacities of a USV, and thus, necessitate careful consideration when formulating the reference trajectory for the USV to adhere to.

We consider that the USV operates in a workspace $\mathcal{W}\subset \mathbb{R}^2$ with obstacles occupying a closed set $\mathcal{O}\subset \mathbb{R}^2$. We define the free space as the open set $\mathcal{A}_{free}:=\left \{  x_1 \in \mathbb{X}: \mathcal{A}(x_1)\cap \mathcal{O}=\emptyset \right \}$, where $\mathcal{A}(x_1)\subset \mathbb{R}^2$ denotes the set that contains the volume of the USV at state $x=x_1$. 
Furthermore, we define the extended free space $\mathcal{A}_{free}(\bar{\rho}):=\left \{  x_1 \in \mathbb{X}: \mathcal{A}(x_1) \cap (\mathcal{O}\oplus \mathcal{B}_{\bar{\rho}})=\emptyset \right \}$, where $\mathcal{B}_{\bar{\rho}}$ is a closed ball of radius $\bar{\rho}$ centered at the origin in $\mathbb{R}^2$, $\bar{\rho}\in \mathbb{R}_{>0}$ is a user-defined distance to the obstacles and $\oplus$ denotes Minkowski addition. 
By “extended” we mean that the USV has some clearance to the obstacles.
The problem we address is the following:
\begin{problem}
    Given the initial state $x(0)\in \mathcal{A}_{free}(\bar{\rho})$ and the goal state $x_g\in \mathcal{A}_{free}(\bar{\rho})$ of the USV, the constraints on the velocity $\mathrm{v}_{\textup{max}}$ and acceleration $\mathrm{a}_{\textup{max}}$, design a reference motion trajectory $p_{\textup{des}} : [0,t_f]\rightarrow \mathbb{R}^2$, for some finite $t_f>0$, and the control laws $F_T$ and $\alpha_r$ 
    that fulfill the control input constraints $F_T \in [0,\bar{F}_T]$, $\alpha_r \in [-\bar{\alpha}_r,\bar{\alpha}_r]$ and the solution $x^*(t)$ of \eqref{eq:model} satisfies $x^*(t)\in\mathcal{A}_{free}$ for all $t\in[0,t_f]$, and $x^*(t_f)=x_g$.\label{prob:setup}
\end{problem}
To solve Problem~\ref{prob:setup}, the following feasibility assumption  is necessary:
\begin{ass}
    There exists an at least twice differentiable trajectory $p_{\textup{des}}:[0,\sigma]\rightarrow \mathcal{P}_{free}(\bar{\rho})$, where $\mathcal{P}_{free}(\bar{\rho}) := \textup{Proj}(\mathcal{A}_{free}(\bar{\rho}))$, is a projection $\textup{Proj}(\cdot) : \mathbb{X} \rightarrow \mathbb{R}^2$ of the set $\mathbb{X}$ onto $\mathbb{R}^2$, with bounded first and second derivatives, i.e. $\| \dot{p}_{\textup{des}}\|\leq \mathrm{v}_{\textup{max}}$, $\| \ddot{p}_{\textup{des}}\|\leq \mathrm{a}_{\textup{max}}$,  such that $p_{\textup{des}}(0)=p(0)$ and $p_{\textup{des}}(\sigma)=p_g$ $= \textup{Proj}(x_g)$.\label{ass:p_des}
\end{ass}

\section{Main Results}
The proposed solution for Problem~\ref{prob:setup} follows a two-layer approach, similar to \cite{verginis2022kdf}, and additionally considers the control input constraints and trajectory optimization. The lower layer is tasked with robust trajectory-tracking control, while the higher level generates a trajectory with B-splines from the path obtained with sampling-based motion planning.

First, we consider the lower layer and the control problem of tracking a given trajectory within user-prescribed bounds. 

\begin{figure}[h!]
\centering
    \includegraphics[width=0.8\linewidth]{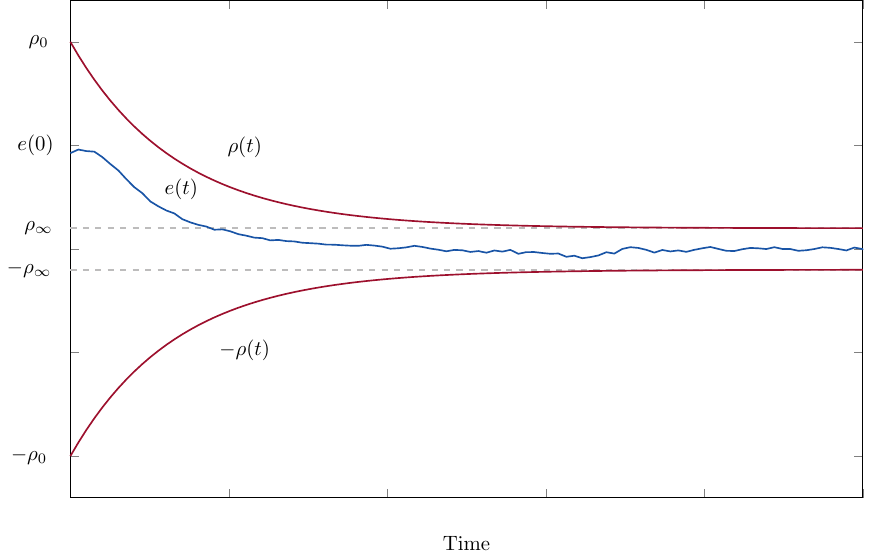} 
    \vspace{-0.3cm}
    \caption{The control objective is that the error evolves inside the prescribed performance funnel.}
    \label{fig:ppc_explanation}    
\end{figure}

The time-varying desired reference trajectory $p_{\textup{des}} = [p_{x,\textup{des}},p_{y,\textup{des}}]^T:[0,\infty) \to \mathbb{R}^2$ will be obtained from the higher layer. For now,
$p_{\textup{des}}$ are assumed  at least twice continuously differentiable functions of time, with bounded first and second derivatives.  
Prescribed performance dictates that a tracking error signal evolves strictly within a funnel
\begin{equation*}
    -\rho(t) < e(t) < \rho(t), \quad \forall t \geq 0,
\end{equation*}
defined by prescribed, exponentially decaying functions of time $\rho(t)$
\begin{equation}
    \rho(t) = (\rho_{0} - \rho_{\infty})e^{-l t} + \rho_{\infty}, \quad \forall t \geq 0, \label{eq:performance_fcn}
\end{equation}
with positive chosen constants $\rho_{0},\rho_{\infty}>0$ and $l\geq 0$ as depicted in Fig.~\ref{fig:ppc_explanation}, thus achieving desired performance specifications, such as maximum overshoot, convergence speed, and maximum steady-state error.

However, as mentioned before, the USV model \eqref{eq:model} is underactuated, and hence the original PPC methodology cannot be directly applied \cite{lapandic2022robust,verginis2022robust}. Consequently, we modify the PPC methodology to achieve trajectory tracking with prescribed performance for the position. 
\begin{figure}[t]
\centering
    \includegraphics[width=\columnwidth]{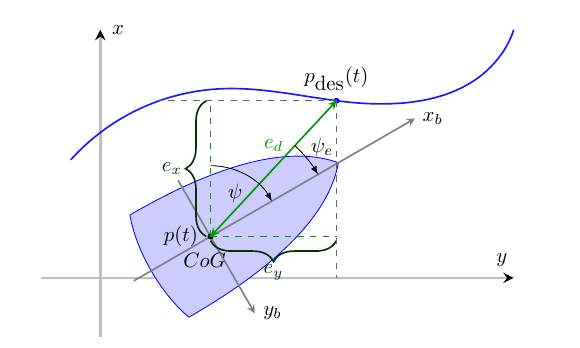} 
    \caption{The considered transformation in NED inertial frame.}
    \label{fig:sketch}  
\end{figure}
 This can be achieved by introducing the distance error $e_d$ and orientation error $e_o$ as
 \begin{subequations}\label{eq:error_d_o_transform}
\begin{align}
    e_d &= \sqrt{e_x^2+e_y^2}, \label{eq:e_d_def} \\
    e_o &= \frac{e_x}{e_d}\sin \psi - \frac{e_y}{e_d}\cos\psi = \sin \psi_e \label{eq:e_o_def}
\end{align}%
\end{subequations}%
where $\psi_e$ is the angle between $\tilde{e}_d=[e_x, e_y]^T$ and the orientation vector $o = [\cos\psi, \sin\psi]^T$ that is defined as the unit vector representing the orientation. 
Note that the orientation error $e_o$, as well as its derivative, only exist if $e_d \neq 0$.

The issue of $e_d$ taking the value zero is not unique to our work but is a common challenge in the control of planar motion of physical systems. This problem is inherent in the nonholonomic nature of these systems that use a transformation of the relative position in the inertial world frame to the distance error $e_d$ and the orientation error in some points ($e_d=0$) is not defined \cite{indiveri1999kinematic,breivik2008guidance, bechlioulis2016trajectory,verginis2015decentralized}. 
This is addressed by designing the funnel boundaries that will always preserve $e_d>\rho_{d,\textup{min}}>0$, where $\rho_{d,\textup{min}}$ is a positive constant, and therefore, the orientation error is well-defined at all times.
Moreover, due to \eqref{eq:e_o_def} it is reasonable to select the funnel boundaries on the orientation error $e_o$ such that $|e_o|<1$. 
The errors between the given reference $p_{\textup{des}}\in\mathbb{R}^2$ and position $p=[p_x,p_y]^T$ in NED (North-East-Down) inertial frame as depicted in Figure~\ref{fig:sketch} are
\begin{align*}
    e_x &= p_{x,\textup{des}} - p_x, \\
    e_y &= p_{y,\textup{des}} - p_y.
\end{align*}

The control objective is to guarantee that the distance and orientation errors evolve strictly within a funnel dictated by the corresponding exponential performance functions $\rho_{d}(t)$, $\rho_{o}(t)$, which is formulated as 
 \begin{subequations}\label{eq:ppc objective}
\begin{align}
    0 < \rho_{d,\textup{min}} &< e_d(t) < \rho_d(t)\\
    -\rho_{o}(t) &< e_{o}(t) < \rho_{o}(t)
\end{align}%
\end{subequations}%
for all $t \geq 0$, given the initial funnel compliance $\rho_{d,\textup{min}} < e_d(0) < \rho_d(0)$, $|e_{o}(0)| < \rho_{o}(0)$. The adopted exponentially-decaying performance functions are 
$\rho_{i}(t) = (\rho_{i,0} - \rho_{i,\infty} )\exp(-l_{i}t) + \rho_{i,\infty}$, $i\in \{d,o\}$. 
Note that, in our setup, it is desired for $e_d$ to not be zero to avoid the singularity and to remain inside the performance funnel whose lower bound is greater than zero. 

Problem~\ref{prob:setup} can be divided into two subproblems. The first subproblem involves the selection of the reference trajectory $p_\textup{des}$, and the second subproblem is to track this trajectory within specified bounds while respecting the input constraints. 
If an algorithm is designed such that the closed-loop system satisfies equations \eqref{eq:ppc objective} and the input constraints are met, then the second subproblem is solved. This ensures that the solution $x^*(t)$ of equation \eqref{eq:model} remains within the free region $\mathcal{A}_{free}$.
The first subproblem, which involves the choice of $p_\textup{des}$, will be addressed in the subsequent sections of the paper.

\subsection{Transformed dynamics}

We use the specific characteristics of the underactuated model of the USV to design prescribed performance controllers. Since $Y$ is proportional to $N$ and we can effectively control only one of them, the idea is to control the forward velocity dynamics $u$ and rotation $r$, while keeping the lateral velocity $v$ below a specified threshold.
Using the introduced transformation \eqref{eq:error_d_o_transform}, the error dynamics are
\begin{subequations}
\begin{align}
    \dot{e}_d &= -u \cos \psi_e + v\sin \psi_e \nonumber \\
    &\phantom{x}+ \dot{p}_{x,\textup{des}} \cos(\psi-\psi_e) + \dot{p}_{y,\textup{des}} \sin(\psi-\psi_e) \\
    \dot{e}_o &= r \cos\psi_e + \frac{u}{e_d} \sin \psi_e \cos \psi_e + \frac{v}{e_d} \cos^2 \psi_e \nonumber\\
    &\phantom{x}- \frac{\dot{p}_{x,\textup{des}}}{e_d}(\sin \psi_e \cos(\psi-\psi_e) - \sin \psi) \\
    &\phantom{x}-\frac{\dot{p}_{y,\textup{des}}}{e_d} (\sin \psi_e \sin(\psi-\psi_e) + \cos \psi) \nonumber\\
     m\dot{u} &= f_u(x,t) + X \\
    m\dot{v} &= -k_v v + f_v(x,t) + c N\label{eq:model_final_v}\\
    I_z\dot{r} &= f_r(x,t) + N 
\end{align}\label{eq:model_final}%
\end{subequations}%
where we used the observation from \eqref{eq:tau_act} that $Y = c N, c = 1/\Delta_x$.

\label{sec:main_results}


\subsection{Control Design}\label{ssec:contro_design}
Now, we describe the proposed control design procedure.
\subsubsection{PPC on distance error} We define the distance normalized error with asymmetric funnel
\begin{equation}
    \xi_d = \frac{2e_d(t)-\rho_d(t) - \rho_{d,\textup{min}}}{\rho_d(t)-\rho_{d,\textup{min}}} \label{eq:first}
\end{equation}
where $\rho_d(t)$ is the prescribed performance function as defined in Eq.~\eqref{eq:performance_fcn} such that $\rho_d(t)>\rho_{d,\textup{min}}>0$ for all $t\geq0$. The asymmetric funnel is chosen to impose the guarantee on $e_d$ to be always positive.
Thus, the distance-error control objective is equivalent to keeping $\xi_d(t)$ inside of $-1<\xi_d(t)<1$, which is equivalent to $0<\rho_{d,\textup{min}}<e_d(t)<\rho_d(t)$, for all $t\geq0$. Hence, $e_o$ and $\dot{e}_o$ will be well defined. Then, we define the transformation
\begin{equation*}
    \varepsilon_d = \mathrm{T}(\xi_d(t))
\end{equation*}
where the transformation $\mathrm{T}$ is a strictly increasing, bijective function $\mathrm{T}: (-1,1) \rightarrow (-\infty,\infty)$ defined as
\begin{equation}
    \mathrm{T}(\xi(t)) =  \textup{atanh}(\xi(t)) =  \frac{1}{2} \ln{\frac{1+\xi(t)}{1-\xi(t)}}.\label{eq:transformation}
\end{equation}
Therefore, maintaining the boundedness of $\mathrm{T}(\xi(t))$ we achieve the control objective of keeping $\xi(t)\in(-1,1)$.
We design the forward velocity reference signal
\begin{equation}
    u_{\textup{des}} = k_d \varepsilon_d \label{eq:u_ref}
\end{equation}
where $k_d>0$ is a user-defined gain.
\subsubsection{PPC on forward velocity error} We define the forward velocity error as
\begin{equation*}
    e_u = u-u_{\textup{des}}.
\end{equation*}
Then, following similar procedure for the error $e_u$, we introduce the performance function $\rho_u(t)$ with $\rho_u(0) > |e_u(0)|$ 
and define the normalized error with a symmetric funnel
\begin{equation}
    \xi_u = \frac{e_u(t)}{\rho_u(t)}. \label{eq:ksi_u}
\end{equation}%
The transformation $\mathrm{T}$ gives us $\varepsilon_u = \mathrm{T}(\xi_u)$. Then, we design the desired generalized force, with a gain $k_u>0$, as
\begin{equation}
    X_{\textup{des}} = -k_u \varepsilon_u \label{eq:X_act}
\end{equation}
\subsubsection{PPC on orientation error} Following a similar procedure as above, we define the normalized error
\begin{equation*}
    \xi_o = \frac{e_o(t)}{\rho_o(t)}
\end{equation*}
with appropriate $\rho_o(t)$, $\rho_o(0)>|e_o(0)|$. We define the transformation $\varepsilon_o = \mathrm{T}(\xi_o(t))$ and design the angular velocity reference 
\begin{equation}
    r_{\textup{des}} = -k_o \varepsilon_o. \label{eq:r_ref}
\end{equation}
where $k_o>0$ is a gain.
\subsubsection{PPC on angular velocity error}
Finally, we define the angular velocity error 
\begin{equation*}
    e_r = r - r_{\textup{des}}, 
\end{equation*}
its normalized version
\begin{equation*}
    \xi_r = \frac{e_r(t)}{\rho_r(t)}
\end{equation*}
and the transformation $\varepsilon_r = \mathrm{T}(\xi_r)$. Then, we design the desired generalized torque, with a gain $k_r>0$, as
\begin{equation}
    N_{\textup{des}} = -k_r \varepsilon_r \label{eq:N_act}.
\end{equation}
Note that, all gains $k_i$, $i=\{d,u,o,r\}$, are user-defined. 

Also note that the desired generalized force $X_{\textup{des}}$ and torque $N_{\textup{des}}$ are realized through the actual control inputs, which are constrained, and thus we examine the input constraints in the next section.
\subsection{Handling input constraints}

In \ref{ssec:contro_design}, we derived the desired generalized force $X_{\textup{des}}$ and torque $N_{\textup{des}}$ that must be realized by the controlled thrust $F_T$ and the controlled rudder position $\alpha_r$ as in \eqref{eq:tau_act}. The desired generalized force \eqref{eq:X_act} and torque \eqref{eq:N_act} are in $\mathbb{R}^2$ while the control inputs are physically constrained to be in the set
\begin{equation}
    (F_T, \alpha_r) \in \mathbb{U} = [0,\bar{F}_T] \times [-\bar{\alpha}_r,\bar{\alpha}_r], \label{eq:input_constraints}
\end{equation}
where $\bar{F}_T>0$ denotes the maximal thrust of the engine and $\bar{\alpha}_r \in (0, \pi/6]$ represents the maximum rudder position. Therefore, an additional mapping is required to obtain the control inputs from \eqref{eq:tau_act}, \eqref{eq:X_act}, and \eqref{eq:N_act}. 

We define the following saturation functions to model these constraints:
\begin{equation}
    \alpha_r = \sigma_{\alpha_r} (\star) = \begin{cases}
       \star, & |\star| \leq \bar{\alpha}_r\\
       \bar{\alpha}_r \sgn (\star), & |\star| > \bar{\alpha}_r
    \end{cases} \label{eq:sat_alpha}
\end{equation}
and
\begin{equation}
    F_T = \sigma_{F_T} (\star) = \begin{cases}
       \bar{F}_T, & \star > \bar{F}_T\\
       \star, & \star \leq \bar{F}_T\\
       0, & \star < 0
    \end{cases}  \label{eq:sat_F}
\end{equation}
Then, the actual applied control $\tau_{act}$ will depend on the saturation functions \eqref{eq:sat_alpha} and \eqref{eq:sat_F}.

Let us introduce two virtual control inputs $u_\alpha, u_F\in \mathbb{R}$ as 
\begin{equation*}
\begin{aligned}[c]
X_{\textup{des}} &= u_F \cos u_\alpha \\
N_{\textup{des}} &= \Delta_x u_F \sin u_\alpha
\end{aligned}
\end{equation*} 
Therefore,
\begin{equation*}
u_\alpha = \arctan \left ( \frac{N_{\textup{des}}}{\Delta_x X_{\textup{des}}} \right ) = \arctan \left ( k_\alpha \frac{\varepsilon_r}{\varepsilon_u} \right )
\end{equation*}
where $k_\alpha = \frac{k_r}{\Delta_x k_u}$, and we set the control input
\begin{equation}
\alpha_r = \sigma_{\alpha_r} (u_\alpha) = \sigma_{\alpha_r} \left(\arctan \left ( k_\alpha \frac{\varepsilon_r}{\varepsilon_u} \right ) \right). \label{eq:saturated_alpha}
\end{equation}
Note that this is only valid when $u_F \neq 0$.
Furthermore,
\begin{equation*}
u_F = \frac{X_{\textup{des}}}{\cos \alpha_r} = -\frac{k_u \varepsilon_u}{\cos \alpha_r}
\end{equation*}
and we set the control input
\begin{equation}
    F_T = \sigma_{F_T} (u_F) = \sigma_{F_T}\left(- \frac{k_u \varepsilon_u}{\cos \alpha_r} \right). \label{eq:saturated_thrust}
\end{equation}
where $\alpha_r$ is implemented according to its definition in \eqref{eq:saturated_alpha}.
Also note that in the last two equations, we immediately used the actual control input $\alpha_r$ rather than the virtual one $u_\alpha$ as it will produce less oscillations to $F_T$ control input when $|u_\alpha|>\bar{\alpha}_r$ is saturated. 

The saturation functions induce an additional block within the closed-loop system and require further stability analysis.  In the next section, we show the conditions on the input constraints under which stability is guaranteed.

\subsection{Stability}\label{ssec:stability}
We present the stability guarantees of the proposed control design under input constraints. 
\begin{thm} \label{thm:main}
Consider the transformed USV dynamics \eqref{eq:model_final} under the proposed control scheme \eqref{eq:first}-\eqref{eq:saturated_thrust}. If the following assumptions hold
\begin{subequations}
\begin{align}
    0&<\munderbar{F}_T  \leq F_T \label{eq:assumption_F_lb} \\
    \bar{F}_u &\leq \bar{F}_T\cos \bar{\alpha}_r \label{eq:assumption_u} \\
    \bar{F}_r &\leq \Delta_x \munderbar{F}_T \sin \bar{\alpha}_r \label{eq:assumption_r}  \\
    |\psi_e (0)| &< \frac{\pi}{2} \label{eq:assumption_psi_e}     
\end{align}
\end{subequations}
where $\munderbar{F}_T$ is a positive constant, $\bar{F}_T$ and $\bar{\alpha}_r$ are the input constraints \eqref{eq:input_constraints}, and $\bar{F}_u$ and $\bar{F}_r$  are appropriate positive constants,
then it holds that $\rho_{d,\textup{min}} < e_{d}(t) < \rho_{d}(t)$, $|e_o(t) | < \rho_o(t) $ 
and all closed-loop signals are bounded for all $t\geq0$.
\end{thm}%
\begin{rem} As mentioned in the problem formulation, the considered surface vehicle can only move forward (without relying on the drag or disturbances), thus $u \geq 0$. 
Consequently, to reach the reference \eqref{eq:u_ref} and reduce the forward velocity $u$, the only options available are to either reduce the forward thrust or completely cut it and rely solely on the drag forces. 
Because the model is unknown and no prediction is used, this behaviour can occur and is an inherent property of the considered underactuated vehicle.  
Suppose one wants to consider the usage of drag forces for braking too. In that case, the actuator model must be augmented with a part that depends on the velocities of the vehicle and surrounding water and the position of the rudder, which is, however, outside of the scope of this work. Therefore, we restrict ourselves only to the case when the applied thrust is positive, as in \eqref{eq:assumption_F_lb}. Future work will include the backward motion, where these effects will be considered. \label{rem:forward_motion}
\end{rem}
\begin{rem}
The assumptions \eqref{eq:assumption_u} and \eqref{eq:assumption_r} can be intuitively explained as the guarantee that for the worst case of the applied generalized force $X$ and torque $N$, there exists enough energy to override the dynamics of the system. 
 When the trajectory evolves inside of the funnels, the constants $\bar{F}_u$ and $\bar{F}_r$ can be understood as the upper bounds on the sum of three components that depend on the dynamics and disturbances, the derivative of the reference signal, in this case $\dot{u}_{\textup{des}}$ and $\dot{r}_{\textup{des}}$, respectively, and a term $l_i(\rho_{i,0} - \rho_{i,\infty})$, $i={u,r}$, that prescribes the speed of convergence.  The constants $\bar{F}_u$ and $\bar{F}_r$ will be explicitly defined in the proof of Theorem 1. 
Please further note that the conditions of Theorem 1 are only sufficient. 
The experiment thus, in a sense, reveals the design flexibility of the approach.
\end{rem}
\begin{rem}
The requirement that $|\psi_e(0)|< \frac{\pi}{2}$ is needed to ensure the initial compliance and boundedness of the orientation error. Moreover, it is shown that $\psi_e$ will remain bounded as $|\psi_e(t)|< \frac{\pi}{2}$, for all $t\geq 0$, which means that $p_{\textup{des}}(t)$ will always be on the positive side of the body axis $x_b$, as seen on Fig.~\ref{fig:sketch}. Thus, the reference trajectory is always kept in front of the USV in its body frame. This is a reasonable behaviour given the observations on the forward motion of the USV and Remark~\ref{rem:forward_motion}.
\end{rem}
\begin{rem}
The presented funnel design for distance error $e_d$ serves as clearance in the motion planner in Section~\ref{sec:motionplanner}, enabling the derivation of a collision-free trajectory. It is crucial to highlight that the choice of the funnel characteristics can be customized by the user. 
However, in practice, selecting excessively small values for $\bar{\rho}$ is not advisable due to the potential generation of large control inputs, which might not be realizable by real actuators. Therefore, one must take into account the capabilities of the system when choosing $\bar{\rho}$ and the funnel functions. 
\end{rem}
\renewcommand\proof{\noindent\hspace{2em}{\itshape Proof of Theorem 1: }}
\begin{proof}
The proof proceeds in two steps. First, we show the existence of a local solution such that $\xi_d(t)$, $\xi_o(t)$, $\xi_u(t)$, $\xi_r(t)$ $\in (-1,1)$, for a time interval $t \in [0,t_{\max})$.  
Next, we show that the proposed control scheme retains the aforementioned normalized signals in compact subsets of $(-1,1)$ in all saturation modes, which leads to $t_{\max} = \infty$, thus completing the proof.

\underbar{\textbf{Part I:}} First, we consider the transformed state vector $\chi=[e_d,e_o,u,v,r]\in \mathcal{X}=\mathbb{R}^5$ that corresponds to the transformed error dynamics in \eqref{eq:model_final} and we define the open set:
\begin{align}
    \Omega =& \big\{ (\chi,t) \in \mathcal{X} \times [0,\infty) : {\xi}_d \in (-1,1), \xi_o \in (-1,1), \notag \\
    &\xi_u \in (-1,1), \xi_r \in (-1,1),  |v| < \bar{v} \big\},
\end{align}
where $\bar{v}$ is a positive constant that will be given later and is only relevant for analysis purposes. 

Note that the choice of the performance functions at $t=0$ implies that ${\xi}_d(0)$, ${\xi}_o(0)$, ${\xi}_u(0)$, ${\xi}_r(0)$ $\in (-1,1)$, implying that $\Omega$ is nonempty. By combining \eqref{eq:model_final}, \eqref{eq:saturated_alpha}, and \eqref{eq:saturated_thrust}, we obtain the closed-loop system dynamics $\dot{\chi} = f_\chi(\chi,t)$, where $f_\chi: \mathcal{X} \times [0,t_{\max})$ is a function continuous in $t$ and locally Lipschitz in $\chi$. Therefore, the conditions of Theorems 2.1.1(i) and 2.13 in \cite{bressan2007introduction} are satisfied, and we conclude that there exists a unique and local solution $\chi:[0,t_{\max}) \to \mathcal{X}$ such that $({\chi}(t),t) \in \Omega$ for all $t \in [0,t_{\max})$. Therefore, it holds that 
\begin{subequations} \label{eq:ksi local bound}
\begin{align}
    {\xi}_d &\in (-1,1) \\
    {\xi}_o &\in (-1,1) \label{eq:ksi_local_o} \\
    {\xi}_u &\in (-1,1) \\
    \xi_r &\in (-1,1) \\
     v    &\in (-\bar{v},\bar{v} ) 
\end{align}
\end{subequations}
for all $t \in [0,t_{\max})$. We next show that the normalized errors in \eqref{eq:ksi local bound} remain in compact subsets of $(-1,1)$ and $(-\bar{v},\bar{v} )$. Note that  \eqref{eq:ksi local bound} implies that that transformed errors ${\varepsilon}_d$, ${\varepsilon}_o$, ${\varepsilon}_u$, $\varepsilon_r$, are well-defined for $t \in [0,t_{\max})$. 

\underbar{\textbf{Part II:}} 
Consider now the candidate Lyapunov function 
\begin{equation*}
    V_d = \frac{1}{4}\varepsilon_d^2
\end{equation*}
Differentiating $V_d$ along the local solution $\chi(t)$ of the reduced error dynamics \eqref{eq:model_final} we obtain
\begin{align*}
    \dot{V}_d &= \varepsilon_d s_d (\rho_d-\rho_{d,\textup{min}})^{-1}(\dot{e}_d-\frac{\dot{\rho}_d}{2}(1+\xi_d) ) \\
    &= \varepsilon_d s_d (\rho_d-\rho_{d,\textup{min}})^{-1} ( -u \cos \psi_e + v\sin \psi_e \nonumber \\
    & \phantom{x}+ \dot{p}_{x,\textup{des}} \cos(\psi-\psi_e) + \dot{p}_{y,\textup{des}} \sin(\psi-\psi_e)-\frac{\dot{\rho}_d}{2}(1+\xi_d) )
\end{align*}
where $s_d = \frac{\mathrm{d}}{\mathrm{d}t}T(\xi)=\frac{1}{1-\xi_d^2}$.
Because of \eqref{eq:ksi_local_o}, it holds that $|e_o(t)| = |\sin(\psi_e(t))| < \rho_o(t) \leq \rho_o(0) < 1$ for all $t\in [0,t_{\max})$.
That means that $|\psi_e(t)|  \leq \arcsin(\rho_o(0)) < \frac{\pi}{2}$
and hence $|\cos(\psi_e(t))| \geq \cos(\arcsin(\rho_o(0))) = \munderbar{c}_{\psi_e} > 0$ for all $t\in[0,t_{\max})$.

Using $u = u_{\textup{des}}+e_u$, \eqref{eq:u_ref}, the boundedness of $\dot{p}_{\textup{des}},\rho_d$ and \eqref{eq:ksi local bound} we obtain 
\begin{equation*}
    \dot{V}_d \leq -k_d \munderbar{c}_{\psi_{e}}|s_d (\rho_d-\rho_{d,\textup{min}})^{-1}|  \varepsilon_d^2   + |s_d (\rho_d-\rho_{d,\textup{min}})^{-1}| |\varepsilon_d|\bar{F}_d
\end{equation*}
where $\bar{F}_d$ is a constant, independent of $t_{\max}$, satisfying
\begin{equation*}
\begin{split}
    \bar{F}_d \geq |& e_u \cos \psi_e -v\sin \psi_e +\dot{p}_{x,\textup{des}} \cos(\psi-\psi_e)    \\
    & - \dot{p}_{y,\textup{des}} \sin(\psi-\psi_e) -\frac{\dot{\rho}_d}{2}(1+\xi_d) |.
\end{split}
\end{equation*}
for all $t \in [0,t_{\max})$.  
This shows the ultimate boundedness of $\varepsilon_d$, i.e. $\dot{V}_d < 0$ when $\frac{\bar{F}_d}{k_d}<|\varepsilon_d|$. Thus $\varepsilon_d$ is ultimately bounded by Theorem 4.18 of \cite{Khalil_nonlinear} as
\begin{subequations}\label{eq:bound_ksi_d}
\begin{equation}
    \left | \varepsilon_d \right | \leq \bar{\varepsilon}_d = \max \left \{ \left | \varepsilon_d (0)  \right |, \frac{\bar{F}_d}{k_d\munderbar{c}_{\psi_{e}}} \right \} \label{eq:epsilon_d_bound}
\end{equation}
for all $t \in [0,t_{\max})$. By employing the inverse of \eqref{eq:transformation}, we obtain
\begin{equation}  
 |{\xi}_{d}(t)| \leq \bar{\xi}_d = \tanh{\bar{\varepsilon}_d} < 1 
\end{equation}
\end{subequations}
and $0<\rho_{d,\textup{min}}<e_d(t)<\rho(t)$, for all $t \in [0,t_{\max})$.

Following the same procedure for $e_o$ and considering a candidate Lyapunov function $V_o = \frac{1}{2}\varepsilon_o^2$, we differentiate $V_o$ along the local solution $\chi(t)$ of the reduced error dynamics \eqref{eq:model_final} to obtain
\begin{equation*}
\begin{split}
    \dot{V}_o = &\varepsilon_o s_o \rho_o^{-1}(\dot{e}_o-\dot{\rho}_o\xi_o)\\
    =&\varepsilon_o s_o \rho_o^{-1}( r \cos\psi_e  + \frac{u}{e_d} \sin \psi_e \cos \psi_e + \frac{v}{e_d} \cos^2 \psi_e  \\
    &- \frac{\dot{p}_{x,\textup{des}}}{e_d}(\sin \psi_e \cos(\psi-\psi_e) - \sin \psi)  \\
    &-\frac{\dot{p}_{y,\textup{des}}}{e_d} (\sin \psi_e \sin(\psi-\psi_e) + \cos \psi) - \dot{\rho}_o\xi_o).
\end{split}
\end{equation*}
Using $r = r_{\textup{des}}+e_r$, \eqref{eq:r_ref}, the boundedness of $\dot{p}_{\textup{des}},\rho_o$ and \eqref{eq:ksi local bound} we obtain 
\begin{equation*}
    \dot{V}_o \leq -k_o \munderbar{c}_{\psi_e}|s_o \rho_o^{-1}| \varepsilon_o^2  + |s_o \rho_o^{-1}\varepsilon_o|\bar{F}_o
\end{equation*}
where 
\begin{equation*}
\begin{split}
    \bar{F}_o \geq & | e_r \cos\psi_e  + \frac{u}{e_d} \sin \psi_e \cos \psi_e + \frac{v}{e_d} \cos^2 \psi_e  \\
    &- \frac{\dot{p}_{x,\textup{des}}}{e_d}(\sin \psi_e \cos(\psi-\psi_e) - \sin \psi)  \\
    &-\frac{\dot{p}_{y,\textup{des}}}{e_d} (\sin \psi_e \sin(\psi-\psi_e) + \cos \psi) - \dot{\rho}_o\xi_o|.
\end{split}
\end{equation*}
for all $t \in [0,t_{\max})$. Thus, from the derivative of $V_o$ we can deduce the ultimate boundedness of $\varepsilon_o$ as
\begin{subequations}\label{eq:bound_ksi_o}
\begin{equation}
    \left | \varepsilon_o \right | \leq \bar{\varepsilon}_o = \max \left \{\left | \varepsilon_o (0) \right |, \frac{\bar{F}_o}{k_o \munderbar{c}_{\psi_e}} \right \}  \label{eq:epsilon_o_bound}
\end{equation}
and we get
\begin{equation}  
 |{\xi}_{o}(t)| \leq \bar{\xi}_o = \tanh{\bar{\varepsilon}_o} < 1.
\end{equation}
\end{subequations}

Differentiating $u_{\textup{des}}$ and $r_{\textup{des}}$ and using their definitions \eqref{eq:u_ref} and \eqref{eq:r_ref}, respectively, and \eqref{eq:epsilon_d_bound} and \eqref{eq:epsilon_o_bound}, we conclude the boundedness of $\dot{u}_{\textup{des}}$ and $\dot{r}_{\textup{des}}$.

Furthermore, consider the candidate Lyapunov function $V_u=\frac{1}{2}m\varepsilon_u^2$. Differentiating we obtain
\begin{align*}
    \dot{V}_u &= m\varepsilon_u s_u \rho_u^{-1} \left (\frac{1}{m}X + \frac{1}{m}f_u(x,t) - \dot{u}_{\textup{des}} - \dot{\rho}_u\xi_u \right)\\
    &= \varepsilon_u s_u \rho_u^{-1} X +\varepsilon_u s_u \rho_u^{-1} \left ( f_u(x,t) - m(\dot{u}_{\textup{des}} +\dot{\rho}_u\xi_u) \right)
\end{align*}
Now, $X=F_T\cos\alpha_r$ from \eqref{eq:tau_act} where $F_T$ and $\alpha_r$ as in \eqref{eq:saturated_thrust} and \eqref{eq:saturated_alpha}. Therefore, we must consider several cases for the different saturation levels. Since $\cos \alpha_r \in [ \cos \bar{\alpha}_r,1 ]$ and because $\alpha_r \leq \frac{\pi}{6}$, we have $\cos \alpha_r>0$.
Let us now consider the saturation effects on the thrust that come from the definition of $F_T$ given by \eqref{eq:saturated_thrust} and $\sigma_{F_T}$
\begin{equation*}
    F_T =  \begin{cases}
       \bar{F}_T, & \text{for } \varepsilon_u  <  -\frac{1}{k_u} \bar{F}_T\cos \alpha_r \\
        -\frac{k_u \varepsilon_u}{\cos \alpha_r} , & \text{for } -\frac{1}{k_u} \bar{F}_T\cos \alpha_r \leq \varepsilon_u < 0\\
       0, & \text{for } \varepsilon_u \geq 0
    \end{cases} 
\end{equation*}
Due to Assumption~\eqref{eq:assumption_F_lb} and the discussion in Remark~\ref{rem:forward_motion} we only consider the case when $\varepsilon_u<0$. However, note that stability is not compromised in the case when $\varepsilon_u \geq 0$.
For $-\frac{1}{k_u} \bar{F}_T\cos \alpha_r~\leq~\varepsilon_u < 0$ we have the unsaturated case in which $X = -k_u\varepsilon_u$, leading to
\begin{equation*}
    \dot{V}_u \leq -k_u |s_u \rho_u^{-1}| \varepsilon_u^2  + |s_u \rho_u^{-1}\varepsilon_u|\bar{F}_u, 
\end{equation*}
where 
\begin{equation}
    \left| f_u(x,t)  - m(\dot{u}_{\textup{des}} + \dot{\rho}_u\xi_u) \right | \leq \bar{F}_u,\label{eq:F_u_def}
\end{equation}for all $t \in [0,t_{\max})$. 
This yields the ultimate boundedness of $\varepsilon_u$ with
\begin{subequations}\label{eq:bound_ksi_u}
\begin{equation}
    \left | \varepsilon_u \right | \leq \bar{\varepsilon}_u = \max \left \{\left | \varepsilon_u (0) \right |, \frac{\bar{F}_u}{k_u} \right \}  \label{eq:eps_u_bound}
\end{equation}
\begin{equation}  
 |{\xi}_{u}(t)| \leq \bar{\xi}_u = \tanh{\bar{\varepsilon}_u} < 1.
\end{equation}
\end{subequations}

For the saturated case, $\varepsilon_u < \tilde{\varepsilon}_u~ =  -\frac{1}{k_u} \bar{F}_T\cos \alpha_r$, the applied force  is $X = \bar{F}_T \cos \alpha_r = - k_u \tilde{\varepsilon}_u$ and then $\dot{V}_u$ becomes
\begin{equation*}
    \dot{V}_u \leq -k_u |s_u \rho_u^{-1}| \varepsilon_u \tilde{\varepsilon}_u + |s_u \rho_u^{-1}\varepsilon_u|\bar{F}_u
\end{equation*}
Since $\varepsilon_u < \tilde{\varepsilon}_u < 0$, then
\begin{align*}
        \dot{V}_u &\leq -k_u |s_u \rho_u^{-1}| \varepsilon_u \tilde{\varepsilon}_u + |s_u \rho_u^{-1}\varepsilon_u|\bar{F}_u \\
          &= -k_u |s_u \rho_u^{-1}|(- |\varepsilon_u|) \tilde{\varepsilon}_u + |s_u \rho_u^{-1}\varepsilon_u|\bar{F}_u \\
          &= |s_u \rho_u^{-1}\varepsilon_u|(k_u\tilde{\varepsilon}_u + \bar{F}_u) \leq 0
\end{align*}
because of Assumption~\eqref{eq:assumption_u}, i.e., $\bar{F}_u \leq \bar{F}_T \cos \bar{\alpha}_r \leq \bar{F}_T\cos \alpha_r  = -k_u\tilde{\varepsilon}_u $. The Lyapunov function is negative during the saturation of the thrust, and $|\varepsilon_u(t)|$ and $|\xi_u(t)|$ remain upper bounded as 
\begin{equation*}
    | \varepsilon_u | \leq |\tilde{\varepsilon}_u|, \quad
    |{\xi}_{u}(t)| \leq \tilde{\xi}_u = \tanh{\tilde{\varepsilon}_u} < 1.
\end{equation*}
for all $t\in[0,\tau_{max})$.

Finally, following a similar procedure with $V_r=\frac{1}{2}I_z\varepsilon_r^2$, we obtain that
\begin{equation*}
    \dot{V}_r = \varepsilon_r s_r \rho_r^{-1}N + \varepsilon_r s_r \rho_r^{-1} \left ( f_r(x,t) - I_z(\dot{r}_{\textup{des}} + \dot{\rho}_u\xi_u ) \right )
\end{equation*}
where $N=\Delta_x F_T\sin\alpha_r$. Thus, using \eqref{eq:saturated_alpha} and the definition of $\sigma_{\alpha_r}$, we consider the saturation effects on the rudder angle
\begin{equation*}
    \alpha_r = \begin{cases}
       \arctan \left ( k_\alpha \frac{\varepsilon_r}{\varepsilon_u} \right ) , & \text{for } |\arctan \left ( k_\alpha \frac{\varepsilon_r}{\varepsilon_u} \right )| \leq \bar{\alpha}_r\\
       -\bar{\alpha}_r \sgn (\varepsilon_r), & \text{for } |\arctan \left ( k_\alpha \frac{\varepsilon_r}{\varepsilon_u} \right )| > \bar{\alpha}_r
    \end{cases} 
\end{equation*}
where we used the fact that $\arctan(\star)$ is an odd function and that $\varepsilon_u$ can only approach zero from the negative side due to \eqref{eq:assumption_F_lb}.
The stability of the unsaturated case can be shown using \eqref{eq:saturated_thrust}, then $N=\Delta_x (-\frac{k_u\varepsilon_u}{\cos\alpha_r})\sin\alpha_r=-\Delta_x k_u\varepsilon_u \tan\alpha_r=-k_r\varepsilon_r$ and
\begin{equation*}
    \dot{V}_r \leq -k_r |s_r \rho_r^{-1}| \varepsilon_r^2  + |s_r \rho_r^{-1}\varepsilon_r|\bar{F}_r, 
\end{equation*}
where 
\begin{equation}
    \left |f_r(x,t) - I_z(\dot{r}_{\textup{des}} + \dot{\rho}_r\xi_r) \right | \leq \bar{F}_r,\label{eq:F_r_def}
\end{equation}
for all $t\in[0,t_{\max})$. This shows the ultimate boundedness of $\varepsilon_r$ with
\begin{subequations}\label{eq:bound_ksi_r}
\begin{equation}
    \left | \varepsilon_r \right | \leq \bar{\varepsilon}_r = \max \left \{\left | \varepsilon_r (0) \right |, \frac{\bar{F}_r}{k_r} \right \}  \label{eq:eps_r_bound}
\end{equation}
\begin{equation}  \label{eq:bound xi r}
 |{\xi}_{r}(t)| \leq \bar{\xi}_r = \tanh{\bar{\varepsilon}_r} < 1.
\end{equation} 
\end{subequations}
Note that in the case when $F_T$ is saturated, i.e., $F_T=\bar{F}_T=-\frac{k_u\tilde{\varepsilon}_u}{\cos\alpha_r}$, the result is similar.
For the saturated case it holds that $N= -\Delta_x F_T \sin \bar{\alpha}_r \sgn(\varepsilon_r)$, then $\dot{V}_r$ becomes
\begin{align*}
    \dot{V}_r \leq& |s_r \rho_r^{-1}|( - \varepsilon_r \Delta_x F_T \sin \bar{\alpha}_r \sgn(\varepsilon_r) + |s_r \rho_r^{-1}\varepsilon_r|\bar{F}_r )\\
    &= |s_r \rho_r^{-1}| (-|\varepsilon_r | \sin\bar{\alpha}_r \Delta_x F_T + |\varepsilon_r |\bar{F}_r) \\
    &= -|s_r \rho_r^{-1}\varepsilon_r|(\Delta_x F_T\sin\bar{\alpha}_r - \bar{F}_r)
\end{align*}
and due to Assumptions~\eqref{eq:assumption_F_lb}~and~\eqref{eq:assumption_r}, i.e., $\bar{F}_r \leq \Delta_x \munderbar{F}_T\sin\bar{\alpha}_r \leq \Delta_x F_T\sin\bar{\alpha}_r$, $\dot{V}_r \leq 0$ and $\varepsilon_r$ is asymptotically stable during the saturated period and $|\varepsilon_r(t)|$ and $|\xi_r(t)|$ remain bounded.

To show that the velocity $v$ remains within a compact subset of $(-\bar{v},\bar{v} )$, let us consider the following candidate Lyapunov function $V_v = \frac{1}{2}mv^2$ and differentiate $V_v$ along the local solution $\chi(t)$ of the reduced error dynamics \eqref{eq:model_final} to obtain
\begin{align*}
    \dot{V}_v &= v(-k_v v + f_v(x,t) - ck_r\varepsilon_r) \\
    &\leq -k_v v^2 + \bar{F}_v|v|
\end{align*}
where $\bar{F}_v \geq | f_v(x,t) - ck_r\varepsilon_r|$, for all $t\in[0,t_{\max})$, due to the boundedness of $\varepsilon_r$, shown in \eqref{eq:eps_r_bound}, and $f_v(x,t)$, as introduced in Sec.~\ref{sec:prob_statement} is continuous and uniformly bounded, thus remaining within a compact subset of $\Omega$ it is bounded as well. Thus, we obtain that $v$ remains in a compact set
\begin{equation}
    |v|\leq \bar{v}' = \max \left \{ |v(0)|, \frac{\bar{F}_v}{k_v}\right \}. \label{eq:v_bounded}
\end{equation}
Furthermore, by choosing sufficiently small control gains $k_i$, $i\in\{d,o,u,r\}$, and initial values of the respective performance prescribed functions $\rho_i(0)$, it can be verified that $|v|\leq \bar{v}' < \bar{v}$ which guarantees that $(\chi,t)$ remains in a compact subset of $\Omega$.

What remains to be shown is $t_{\max} = \infty$. To this end, note that \eqref{eq:bound_ksi_d}, \eqref{eq:bound_ksi_o}, \eqref{eq:bound_ksi_u}, \eqref{eq:bound_ksi_r} and \eqref{eq:v_bounded} imply that $(\chi,t)$ remain in a compact subset of $\Omega$, i.e., there exists a positive constant $\munderbar{d}$ such that $d_\mathcal{S}((\chi,t), \partial \Omega) \geq \munderbar{d} > 0$, for all $t \in [0,t_{\max})$,  where $d_\mathcal{S}((\chi,t),\partial \Omega)=\inf\limits_{y\in \partial \Omega} \left \| (\chi,t)-y \right \|$ is the distance of a point $(\chi,t)\in \mathcal{X}\times[0,\infty)$ to the boundary $\partial \Omega$ of the set $\Omega$. Since all relevant closed-loop signals have already been proven bounded, it holds that 
$\lim_{t \to t_{\max}^-} \left( \|\chi(t)\| + d_\mathcal{S}((\chi(t),t), \partial \Omega)^{-1} \right) \leq \bar{d}$, for some finite constant $\bar{d}$, and hence direct application of Theorem~2.1.4 of \cite{bressan2007introduction} dictates that $t_{\max} = \infty$, which completes the proof.\hfill\openbox
\end{proof}
\section{Kinodynamic motion-planning}\label{sec:motionplanner}
In the previous section, we presented a control protocol for the lower layer of the proposed solution for Problem~\ref{prob:setup}.
In this part, we present an optimization-based kinodynamic motion planning algorithm for generating smooth trajectories $p_{\textup{des}}$ that fulfill imposed kinodynamic constraints such as the velocity $\mathrm{v}_{\textup{max}}$ and acceleration $\mathrm{a}_{\textup{max}}$ constraints from Problem~\ref{prob:setup}.
Our approach builds on KDF \cite{verginis2022kdf}, which is a framework that uses PPC to track and keep an agent inside of a funnel around a trajectory obtained from a sampling-based motion planner generated path.
The presented control design from the previous section provides us with the guarantee that the vehicle is going to stay inside of the prescribed funnel bounds around the reference trajectory. 
In this section, we consider cubic B-splines\cite{deboor2001splines} and use some of their properties to generate the reference trajectory that aims to fulfill Assumption~\ref{ass:p_des}. 

First, we sample the extended free space $\mathcal{A}_{free}(\bar{\rho})$ and obtain a path from the initial position to the goal position using Rapidly-exploring Random Trees (RRT) \cite{lavalle2006planning}. The path is used to generate the desired trajectory $p_{\textup{des}}$ in $\mathcal{A}_{free}(\bar{\rho})$ and by employing the described control design that ensures $e_d < \rho_{d,0}$, where $\rho_{d,0} < \bar{\rho}$, the USV maintains a collision-free trajectory, as it remains $\bar{\rho}$-close to $p_{\textup{des}}$.

The RRT obtained path is non-smooth, and smoothening it by interpolating through the path points without considering the obstacles might result in collisions \cite{verginis2022kdf,eshtehardian2022continuous}. 
Therefore, we pose the smoothening problem of the RRT obtained path $\{X_k\}_{k=0}^{N_X-1}$, $X_k\in\mathbb{R}^2$, where $N_X$ is the number of obtained path points, as an optimization problem. The uniform B-splines are determined by $N=N_X+4$ control points $\{ q_k \}_{k=0}^{N-1}$, $q_k\in\mathbb{R}^2$, and $M=N+d+1$ uniformly separated knots $t_k = k \Delta t$, $k=4,...,N$, where $d=3$ is the degree of the curve. The knot spacing, denoted by $\Delta t$, can be set \textit{a priori}, which may result in sub-optimal trajectories or, as in our case, treated as an optimization variable at the expense of computation time. The rest of the knots are defined as $t_0=t_1=t_2=t_3=0$ and $t_N=t_{N+1}=t_{N+2}=t_{N+3}=t_{N+4}$, to guarantee zero velocity and acceleration at the initial and final position imposed with $q_k=X_0$, for $k=0,1,2$ and $q_k=X_{N_X}$, for $k=N-3, N-2, N-1$. The goal is to smoothen the desired trajectory $p_{\textup{des}}$ based on the obtained RRT path. This is achieved by minimizing the distance of the B-spline $p_{\textup{des}}(t_k)$ at each knot from the RRT path points. Using the properties for uniform B-splines, $p_{\textup{des}}$ can be evaluated on a segment $t \in [t_k, t_{k+1})$ with the knowledge of any four consecutive control points $Q_k=[q_{k-3},q_{k-2},q_{k-1},q_{k}]^T$, as
\begin{equation*}
    p_{\textup{des}}(t) = \boldsymbol{u}^TMQ_k
\end{equation*}
where $\boldsymbol{u}= [1, u, u^2, u^3]^T$ is the basis vector with $u=\frac{t - t_k}{t_{k+1}-t_k}\in[0,1)$. The matrix $M$ is a fixed known matrix \cite{qin1998general} independent of $k$ given by
\begin{equation*}
{\small
    M = \frac{1}{6}\begin{bmatrix}
        1 & 4 & 1 & 0 \\
        -3 & 0 & 3 & 0 \\
        3 & -6 & 3 & 0 \\
        -1 & 3 & -3 & 0 
    \end{bmatrix}.}
\end{equation*}
Then the distance at $t_k$ can be calculated as
\begin{equation*}
    \left \| p_{\textup{des}}(t_k) - X_k \right \| = \left \| \boldsymbol{u}^T(0)MQ_k - X_k \right \| = \left \| m^TQ_k - X_k \right \|
\end{equation*}
where $m=\frac{1}{6}[1, 4, 1, 0]^T$.

\renewcommand{\thefigure}{4}
\begin{figure}[t]
\centering
  \includegraphics[width=.70\linewidth]{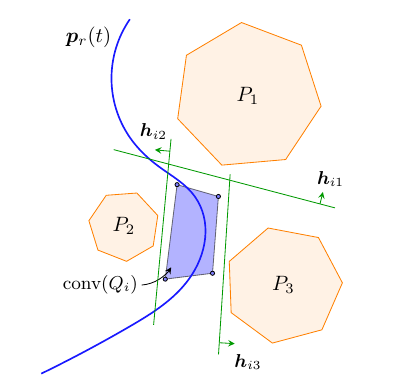}
\caption{The convex hull of a set of four consecutive points is linearly separable from the obstacles. }
\vspace{-0.5cm}
\label{fig:lin_sep_sketch}
\end{figure}

We are also interested in keeping the trajectory inside of the specified velocity $\mathrm{v}_{\textup{max}}$ and acceleration bounds $\mathrm{a}_{\textup{max}}$, such that $\| \dot{p}_{\textup{des}}\|\leq \mathrm{v}_{\textup{max}}$, $\| \ddot{p}_{\textup{des}}\|\leq \mathrm{a}_{\textup{max}}$. Using the derivative properties of B-splines, we can obtain the velocity $\mathrm{v}_k$ and acceleration $\mathrm{a}_k$ at a control point $q_k$ as
\begin{align}
    \mathrm{v}_k &= \frac{d(q_k - q_{k-1})}{t_{k+d}-t_k} = \frac{q_k - q_{k-1}}{\Delta t} \\
    \mathrm{a}_k &= \frac{(d-1)(\mathrm{v}_k - \mathrm{v}_{k-1})}{t_{k+d-1}-t_k} = \frac{\mathrm{v}_k - \mathrm{v}_{k-1}}{\Delta t}
\end{align}
for uniformly spaced knots.
Moreover, it is beneficial to obtain as smooth curve as possible. Thus we include a third derivative (jerk) minimization term in the cost function as 
\begin{equation}
    \left \|j_k  \right \| \Delta t = \frac{1}{\Delta t} \left \| \mathrm{a}_k - \mathrm{a}_{k-1} \right \| \Delta t = \left \| b^T Q_k \right \|
\end{equation}
where $b=[-1,3,-3,1]^T$.
Since we are optimizing over a fixed number of control points and uniformly separated knots, we can also include $\Delta t$ in the optimization problem to minimize the time duration.

Because four consecutive control points create a convex hull around a spline segment, collision avoidance can be done using linear separation. The obstacles are modeled as $n$-sided convex polygons with vertices at $p_{jl}\in\mathbb{R}^2$ of a $j$-th polygon, $l=1,...,n$, compactly written as $P_j=[p_{j1},...p_{jn}]$. The existence of a line that separates the two sets is based on the existence of $\boldsymbol{h}_{ij}\in\mathbb{R}^2$ and $d_{ij}\in\mathbb{R}$, where $i=0,..., N-4$ denotes the $i$-th convex hull around a spline segment determined with $Q_i$, such that 
\begin{align}
    Q_i\boldsymbol{h}_{ij} &> d_{ij}\boldsymbol{1}_{4}, \\
    P_j\boldsymbol{h}_{ij} &< d_{ij}\boldsymbol{1}_{n},
\end{align}
for all $i,j$, where $\boldsymbol{1}_{n}\in\mathbb{R}^n$ denotes a unit vector. The linear separation concept is depicted on Fig.~\ref{fig:lin_sep_sketch}.

\renewcommand{\thefigure}{5}
\begin{figure*}[t!]
\centering
  \includegraphics[width=0.9\textwidth]{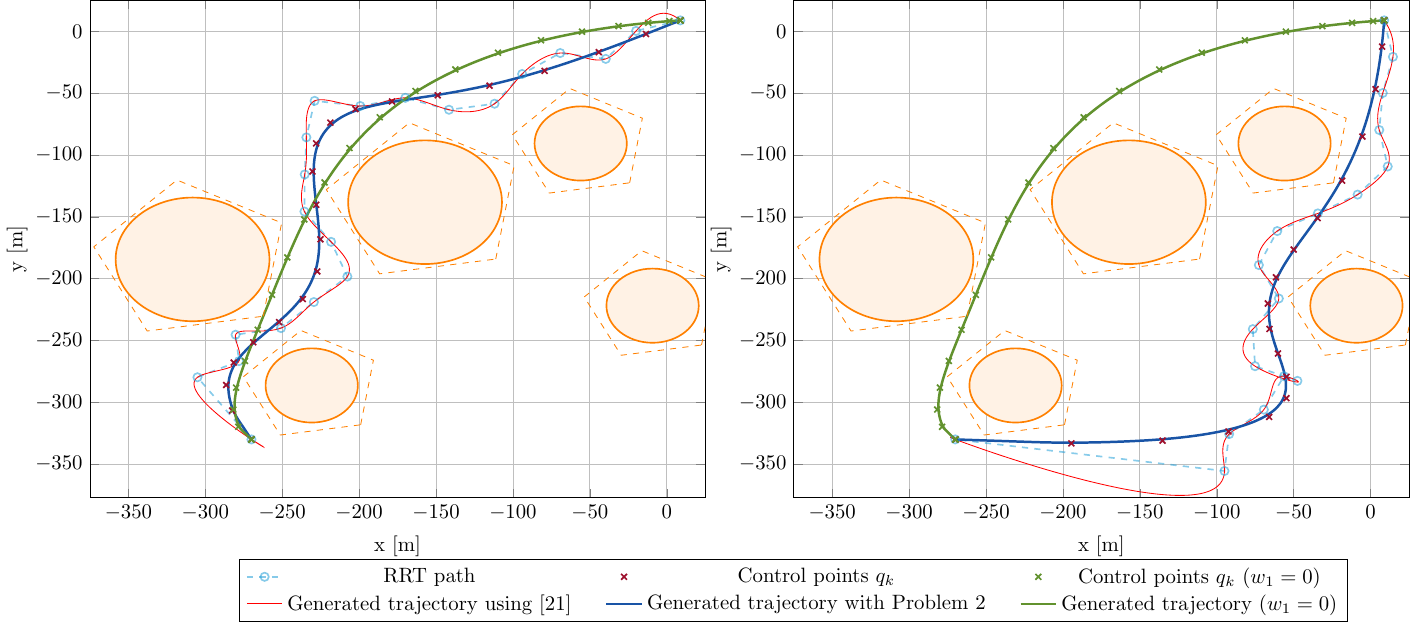}
\caption{The figure shows two runs of the trajectory generation algorithm with RRT and the same setup of obstacles as it will be in the real-world experiments. Two different RRT paths were obtained for comparison. The trajectory is then generated using the optimization problem in Problem~\ref{prob:trajectory_generation}. Moreover, we show the optimization result without RRT points with $w_1 = 0$ in green. Note that interpolating through the RRT points only as in \cite{verginis2022kdf} would result in a difficult trajectory to follow with unnecessary deviations as depicted in red.}
\label{fig:simulation}\vspace{-0.4cm}
\end{figure*}

\renewcommand{\thefigure}{7}
\begin{figure*}[b!]
\centering
  \includegraphics[width=0.9\textwidth]{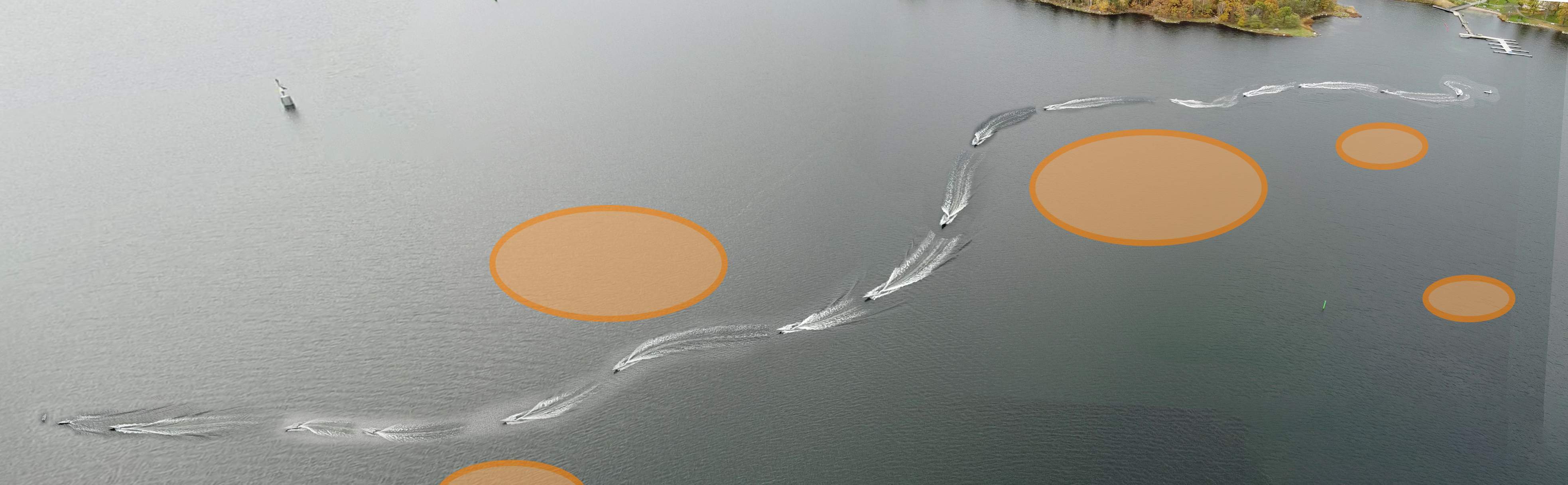}
\caption{The timelapse perspective view (slightly distorted due to the rotation of the UAV camera) of the experimental run with obstacle areas in orange.}
\label{fig:experiment_perspective}\vspace{-0.4cm}
\end{figure*}

Finally, we can state the nonlinear optimization problem:
\begin{problem}[Kinodynamic trajectory generation]
\begin{equation*}
    \min_{\Delta t, q_k, \boldsymbol{h}_{ij},d_{ij}} \sum_{k=3}^{N-4} w_1\left \| m^TQ_{k} -X_{k-2} \right \|^2 + w_2\left \| b^TQ_{k} \right \|^2 + w_3\Delta t 
\end{equation*}
\text{subject to}\label{prob:trajectory_generation}
\begin{subequations}
\begin{alignat}{2}
    &q_k = X_0, \quad  && k \leq 2, \label{eq:init_cnd}  \\
    &q_k = X_{N-5}, \quad  &&k\geq N-3, \label{eq:final_cnd} \\
    &\|q_k - q_{k-1} \| \leq  \mathrm{v}_{\textup{max}}\Delta t , \quad   &&k\geq 1, \label{eq:vel_cstr}\\
    &\|q_k - 2q_{k-1} - q_{k-2}  \| \leq  \mathrm{a}_{\textup{max}}\Delta t^2, \quad &&k\geq2, \label{eq:acc_cstr} \\ 
    &Q_i\boldsymbol{h}_{ij} > d_{ij}\boldsymbol{1}_{4},\quad && \forall i,j \label{eq:lin_sep1}\\
    &P_j\boldsymbol{h}_{ij} < d_{ij}\boldsymbol{1}_{n},\quad && \forall i,j \label{eq:lin_sep2}
\end{alignat}    
\end{subequations}
\end{problem}

In Problem~\ref{prob:trajectory_generation}, the constraints \eqref{eq:init_cnd} and \eqref{eq:final_cnd} denote the initial and final conditions, \eqref{eq:vel_cstr}-\eqref{eq:acc_cstr} the velocity and acceleration constraints, \eqref{eq:lin_sep1}-\eqref{eq:lin_sep2} the linear separability conditions as explained previously. Based on the weight choices $w_i\geq0$, $i=1,2,3$, in Problem \ref{prob:trajectory_generation}, we can prioritize between the three objectives, namely, fitting the curve to the RRT obtained path, minimizing the jerk, and minimizing the time, respectively. Note that we shifted indices of $Q_k$ such that the control points are not optimized over the fixed control points determining the initial and final values of the curve.
Also, note that it is possible to choose $w_1=0$ and thus avoid using the RRT path. However, this option results in an order of magnitude higher execution time, therefore making the RRT path useful prior for the curve generation.

Note that the optimization problem is nonlinear due to \eqref{eq:acc_cstr}. A relaxed optimization problem could be equivalently written as a quadratically constrained quadratic program (QCQP) by omitting the acceleration constraint and some basic mathematical manipulations. 
However, this is out of the scope of this work. 
To solve this optimization problem, we used interior point optimizer (IPOPT) in CasADi \cite{Andersson2019casadi}. 
The computation time depends on how many points are used as the prior from the RRT obtained path, as they directly influence the number of the spline control points which are the variables of the optimization.

A simulation example of the presented trajectory generation algorithm with $\mathrm{v}_{\textup{max}}=10$, $\mathrm{a}_{\textup{max}}=2$ is given on Fig.~\ref{fig:simulation}. 
On average, the optimization process typically completes within a range of 5 to 20 seconds across multiple runs on an  Intel i7-8665U CPU running at 1.90GHz with 8 cores and Ubuntu 18.04 operating system. If the RRT path is not used as the prior in the optimization problem, the optimization might extend to 3-5 minutes on a user-defined number of points.
Furthermore, on Fig~\ref{fig:simulation}, we provide a comparison with trajectories generated using \cite{verginis2022kdf}. 
Note that \cite{verginis2022kdf} uses interpolation through RRT points which might result in collisions with obstacles if the fixed time interval constraint between two points is not chosen appropriately, as the interpolated spline is not checked for collision. 
Furthermore, fluctuations in velocity and acceleration profiles can be significantly reduced with Problem~\ref{prob:trajectory_generation} due to the constraints \eqref{eq:vel_cstr}-\eqref{eq:acc_cstr}.

\section{Experimental Results}\label{sec:experimental_results}

The overall scheme is tested in real-world open-water experiments. 
We used Piraya, an autonomous USV developed by SAAB Kockums AB, as a research platform equipped with cameras, GPS, LIDAR, etc., in the WARA-PS research arena \cite{andersson2021wara}. In this experiment, the GPS is used for determining the absolute position, while the obstacles are defined beforehand and remain static during the experiment.
Piraya is $4$ meters long, and the goal point is approximately $450$ meters away from the initial position with obstacles present. The funnels are chosen to be static ($\rho_0=\rho_\infty$ and $l=0$) during the 3-minute long experiment because the desired error has the same performance criteria throughout the whole experiment. Moreover, they are set relatively loose to avoid saturating the control inputs too often. Thus, $\rho_d = 28$, $\rho_{d,\textup{min}}=0.5$, $\rho_u=25$, $\rho_o = 0.9999$, $\rho_r = 15$.

\renewcommand{\thefigure}{6}
\begin{figure}[h!]
    \centering
    \includegraphics[width=\columnwidth]{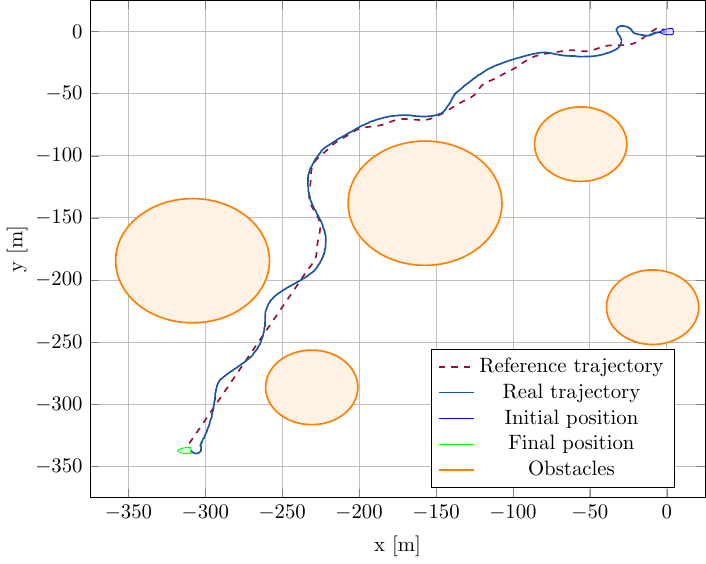}
    \vspace{-0.8cm}
    \caption{Top view of the real-world open water experiments}
        \vspace{-0.7cm}
    \label{fig:kdf2}
\end{figure}

The top view of the experiment is depicted in Figure~\ref{fig:kdf2} while the timelapse perspective view of the experiment is visible in Figure~\ref{fig:experiment_perspective}. We can observe that, neglecting the deviation that occurred in the first moments of the experiment, the surface vehicle was able to follow the trajectory. The initial deviation occurred because the reference errors were relatively small to excite the control algorithm to produce the control thrust $F_T$, so the USV was drifting due to the winds and open water currents. After a few seconds, the control inputs were activated when the errors became larger, and the USV successfully recovered from the deviation. This, however, can be alleviated by tightening the funnels. 

\renewcommand{\thefigure}{\arabic{figure}}
\begin{figure}[t]
    \centering
    \includegraphics[width=\columnwidth]{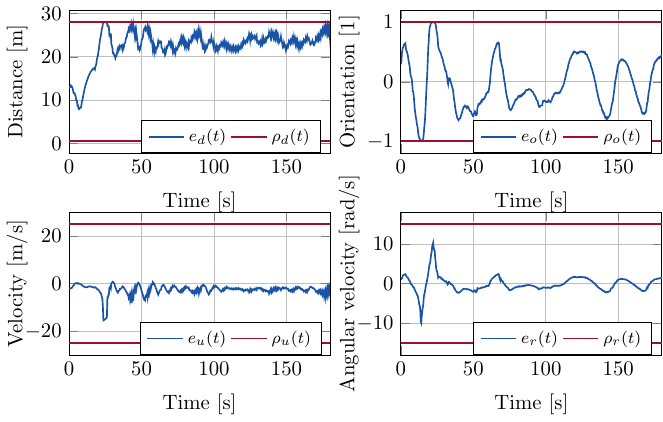}
    \caption{Error evolution and the funnels in the real-world experiment. Note that the orientation error is dimensionless.}
    \label{fig:funnelsss}\vspace{-0.2cm}
\end{figure}

The funnels are depicted in Figure~\ref{fig:funnelsss}, and all signals are bounded as expected. The distance error is always positive, but it is interesting to observe that it is kept relatively close to the funnel boundary during the experiment. This can be explained by the fact that the chosen loose funnel will generate more control thrust when the error $e_d$ is relatively large. 

\begin{figure}[h]
    \centering
    \includegraphics[width=\columnwidth]{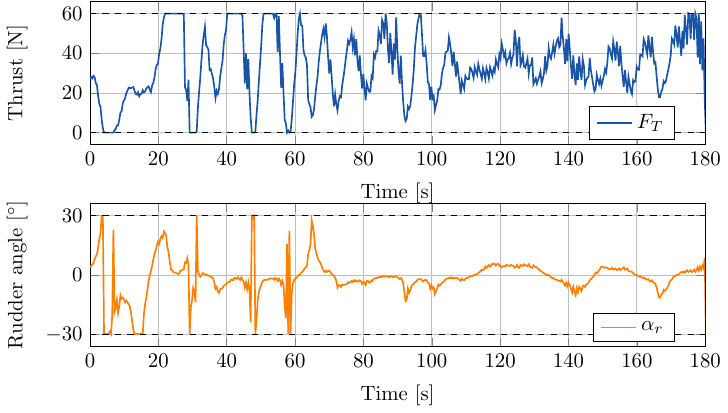}
    \caption{Control inputs in the real-world experiment.}
    \label{fig:inputss}
\end{figure}

In Figure~\ref{fig:inputss}, the control inputs are presented, and periods of saturation are visible. They occur mainly in the first half of the experiment, during which the USV was expected to recover from the initial deviation and subsequent overshoots. The most problematic behaviour is as expected when $F_T=0$, which causes $\alpha_r$ to be saturated, although the orientation error might not require that action. This is discussed in Remark~\ref{rem:forward_motion}. 

\begin{figure}[h]
    \centering
    \includegraphics[width=\columnwidth]{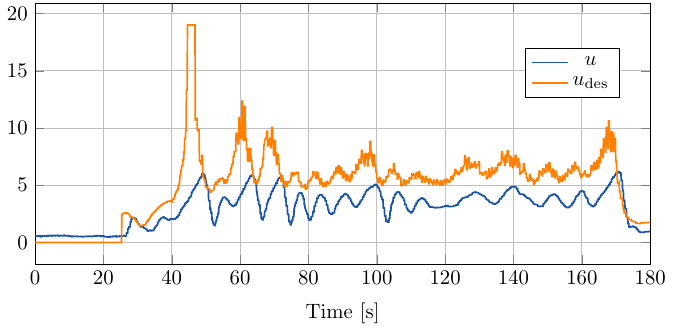}
    \caption{Forward velocity $u$ and its reference $u_{\textup{des}}$ in time. }
    \label{fig:forward_vel_experiments}\vspace{-0.2cm}
\end{figure}

One of the reasons that might cause the unwanted chattering and oscillations in the control input $F_T$ as well as in error $e_d$, visible in Fig.~\ref{fig:funnelsss}, are the unmodeled input delays that are present on the controlled USV. Namely, the engine has inherent delays which may induce these effects, which for a bystander, look like the USV is periodically accelerating and decelerating unnecessarily, which is visible in Figure~\ref{fig:forward_vel_experiments}. Thus, the focus of future work will be on removing these oscillations.


\section{Conclusion}\label{sec:conclusion}
In this paper, we proposed improved kinodynamic motion planning via funnel control (KDF). We derived stability results under input constraints and system underactuation. The presented framework is tested in a real-world open-water experiment. In future work, we plan to investigate how to remove the oscillations in the forward motion that may be due to the control input delays. Moreover, it may be beneficial to explore the effect of the funnel size and performance functions on the behaviour of the vehicle. Furthermore, we plan to redesign the motion planning procedure with B-splines as an iterative online scheme suitable for moving obstacles and dynamic environments.

\section*{Acknowledgments}
This work was partially supported by the Wallenberg AI, Autonomous Systems and Software Program (WASP), the Swedish Science Foundation, the Swedish Research Council, and Knut and Alice Wallenberg Foundation (KAW).

\bibliographystyle{unsrt}
\bibliography{refs.bib}


\end{document}